\documentclass{amia}
\usepackage{lipsum} 
\usepackage{wrapfig}
\usepackage{booktabs}
\usepackage{multirow}
\usepackage{array}
\usepackage{float}
\begin{document}

\title{Local Contrastive Learning for Medical Image Recognition}

\author{Syed A. Rizvi, BS$^1$, Ruixiang Tang, BSC$^2$, Xiaoqian Jiang, PhD$^3$,\\ Xiaotian Ma, MS$^3$, Xia Hu, PhD$^2$ }

\institutes{
    $^1$ Yale University, New Haven, CT; $^2$Rice University, Houston, TX; \\
    $^3$University of Texas Health Science Center, Houston, TX
}

\maketitle

\section*{Abstract}

\textit{The proliferation of Deep Learning (DL)-based methods for radiographic image analysis has created a great demand for expert-labeled radiology data. Recent self-supervised frameworks have alleviated the need for expert labeling by obtaining supervision from associated radiology reports. These frameworks, however, struggle to distinguish the subtle differences between different pathologies in medical images. Additionally, many of them do not provide interpretation between image regions and text, making it difficult for radiologists to assess model predictions. In this work, we propose Local Region Contrastive Learning (LRCLR), a flexible fine-tuning framework that adds layers for significant image region selection as well as cross-modality interaction. Our results on an external validation set of chest x-rays suggest that LRCLR identifies significant local image regions and provides meaningful interpretation against radiology text while improving zero-shot performance on several chest x-ray medical findings.}

\section*{1 \quad Introduction}

Advancements in medical imaging technologies have accelerated the creation of large radiographic image datasets, opening new possibilities for image analysis in patient care. This development, however, has been in parallel with an increasing burden on radiologists to interpret larger numbers of medical images \cite{bhargavan2009workload}. Deep Learning (DL) provides a promising solution for automating the analysis of these large datasets. Deep Neural Networks (DNNs) have demonstrated strong representational learning power at multiple layers of abstraction, without relying on handcrafted features or domain expertise\cite{lecun2015deep}. In the medical domain, DL-based systems have shown strong performance on downstream tasks such as disease classification \cite{esteva2017dermatologist, mckinney2020international} and pathology localization \cite{chen2015standard, ghesu2016artificial, rajpurkar2017chexnet}. Annotating large radiology datasets for training DL models, however, is cost-prohibitive at scale due to the cost and expertise required to annotate medical images.

Self-supervised vision-language frameworks address this challenge by obtaining supervision on images from associated text \cite{radford2021learning}, which comes in the form of associated radiology reports. In Computer Vision, self-supervised Vision-language pretraining frameworks \cite{tsai2019multimodal, chen2020uniter, kim2021vilt, radford2021learning} train on large corpora of paired image and text data, learning alignment between image and text embeddings. Fine-tuning can then be done for downstream tasks such as visual question answering and image-text retrieval \cite{kim2021vilt}. In the medical domain, self-supervised vision-language architectures have shown promising results in classification \cite{yan2022clinical} and medical image-report retrieval \cite{moon2022multi}. Contrastive frameworks \cite{huang2021gloria, zhang2022contrastive, tiu2022expert} in particular have shown strong performance on downstream tasks, including medical image-text retrieval and zero-shot classification, where a model is able to predict new pathologies that were not explicitly annotated in the training set \cite{xian2018zero, tiu2022expert}. Contrastive frameworks operate by encoding images and text into a shared latent space and define an objective that maximizes agreement between paired images and text \cite{chauhan2020joint, chen2020simple}. 

Although powerful, these architectures may struggle to differentiate radiology images, which have less variation between different images than natural images. Medical observations often manifest in small regions of an image, and often many regions of the image contain uninformative background regions which do not contribute meaningful information to the global embedding. Several works have focused on learning both global and local features in input images to emphasize discriminative information contained in local regions. In natural images, TransFG \cite{he2022transfg} analyzed the self-attention maps created by Vision Transformers (ViTs) \cite{dosovitskiy2020image} to identify significant local regions for fine-grained visual recognition tasks. Li et al. \cite{li2019visual} proposed a visual-semantic reasoning model to enhance the relationship between local image regions before contrasting image representations against text data. In radiology image analysis, GLoRIA \cite{huang2021gloria} proposed a local contrastive loss in addition to the normal contrastive objective between the global image and text embeddings. Attention weights are learned between image region embeddings and individual word embeddings, generating context-aware local image embeddings which are then used in a contrastive loss against word embeddings. In the attention-weighted image representations, however, unimportant background image regions are still contributing information to the contextualized local image embedding. Additionally, the frameworks above are standalone architectures, not compatible with other pretrained model architectures.

In addition to downstream performance, the interpretability of computational systems for medical image analysis tasks is crucial for providing context behind model predictions. Contrastive architectures \cite{radford2021learning, chen2020simple, tiu2022expert} gain zero-shot capabilities when pretrained on large amounts of data, however, cannot give reasoning or significant features associated with the prediction. In radiology, computational systems often aid radiologists in analyzing chest radiographs, necessitating a further degree of cooperation between the radiologist and the computational model than a single prediction. Highlighting significant image regions associated with a given text prompt can provide a visual indicator for the radiologist to check the model's prediction and increase radiologist-computer cooperation. In a similar vein, GLoRIA \cite{huang2021gloria} provides attention scores between image regions and text prompts. GLoRIA cannot, however, highlight significant image regions without contrasting all image regions against a text prompt. In addition, the architecture is not immediately compatible with other pretrained contrastive architectures. Additional modularity and flexibility are needed in order to open up possibilities for local region selection and local contrastive learning to other pretrained contrastive models.

In this work, we propose LRCLR, a fine-tuning module aimed at introducing interpretable local region selection while making minimal changes to existing pretrained architectures. LRCLR consists of a region selection module, which selects significant regions in input images solely based on self-attention matrices from the transformer-based image encoder. Selected regions are passed along with text tokens to a cross-modal transformer, which contextualizes significant image regions with texts through self-attention. A local contrastive objective is added to the global objective, contrasting the image and text class tokens output from the cross-modal transformer. LRCLR is compatible with many vision-language architectures as an add-on module during fine-tuning. We evaluate our approach in a zero-shot classification setting on an external dataset of chest x-rays and study the interpretability of rare chest x-ray findings. We demonstrate improved zero-shot classification performance on rare chest x-ray findings and show interpretability results between significant local regions and radiology report text.

\section*{2 \quad Preliminaries}

\subparagraph{Medical Image Recognition}

Medical Image Recognition is a diagnostic tool for clinicians in patient care, involving inspecting multiple modalities of medical image data for patterns of pathology and medical observations \cite{chan2020deep}. With advancements in imaging technologies as well as multi-modal image data being generated in patient examinations, the burden on clinicians has increased to analyze patient information. Initial studies aimed at using computational power to assist in medical image analysis emerged as early as the 1960s, based on image analysis and early Machine Learning (ML) techniques \cite{li2015computer}. The emergence of Deep Learning, paired with increased computational power and availability of medical image data, has accelerated the field of Machine Learning-based medical image analysis, based on the representation learning ability of Deep Neural Networks (DNNs) at multiple levels of abstractions \cite{lecun2015deep}. Recent DL works based on Deep Convolutional Neural Networks (CNNs) \cite{krizhevsky2017imagenet} and Transformers \cite{vaswani2017attention} have shown strong performance on medical image classification \cite{sowrirajan2021moco} and segmentation \cite{cao2023swin} tasks.

\subparagraph{Radiology Data}
Radiology is a branch of medicine that utilizes diagnostic imaging techniques to help treat patients, including x-rays, computed tomography (CT) scans, magnetic resonance images (MRIs), and ultrasounds (US) \cite{ker2017deep}. Common tasks for ML-based systems in radiology include segmentation, disease detection, brain function and neurological activity analysis from fMRIs, image-retrieval systems, and radiology report text analysis \cite{wang2012machine}.

\subparagraph{Self-Supervised Vision-Language Architectures}

Pretraining multimodal vision-language architectures has been shown to improve performance on downstream vision-language tasks. Vision-language frameworks can be divided into several categories, depending on their architectures and encoding strategies. Dual-branch fusion frameworks \cite{tsai2019multimodal} use two parallel encoders for different modalities, with cross-modal attention mechanisms for capturing inter-modal interactions. Single-branch fusion architectures \cite{chen2020uniter, kim2021vilt} define a single sequence-based architecture, usually a transformer, which accepts a sequence of concatenated image and text tokens. The self-attention within the architecture captures cross-modal interactions, with different objectives defined on the model output, such as masked language modeling, masked region modeling, or word-region alignment \cite{chen2020uniter}. Dual encoder-based architectures \cite{radford2021learning} utilize a separate encoder to map images and text into a shared embedding space, where a contrastive objective is used to align their representations. This enables zero-shot capabilities at inference time by introducing new text prompts and images to the trained model.

\subparagraph{Zero-shot Learning}

Zero-shot learning is a paradigm which aims to classify images belonging to classes that were not explicitly labeled in the training dataset. Without training labels, supervision for images is obtained from other data modalities, such as an associated text corpus \cite{karpathy2015deep, socher2013zero, radford2021learning}. By learning alignment between images and text through a contrastive learning objective, new text prompts can be introduced during inference to find the most similar zero-shot label to a given image. Recent works have shown promising results on zero-shot learning in natural images \cite{radford2021learning} and medical image classification tasks \cite{huang2021gloria, tiu2022expert}.

\begin{figure}[t]
  \centering
\includegraphics[width=0.9\textwidth]{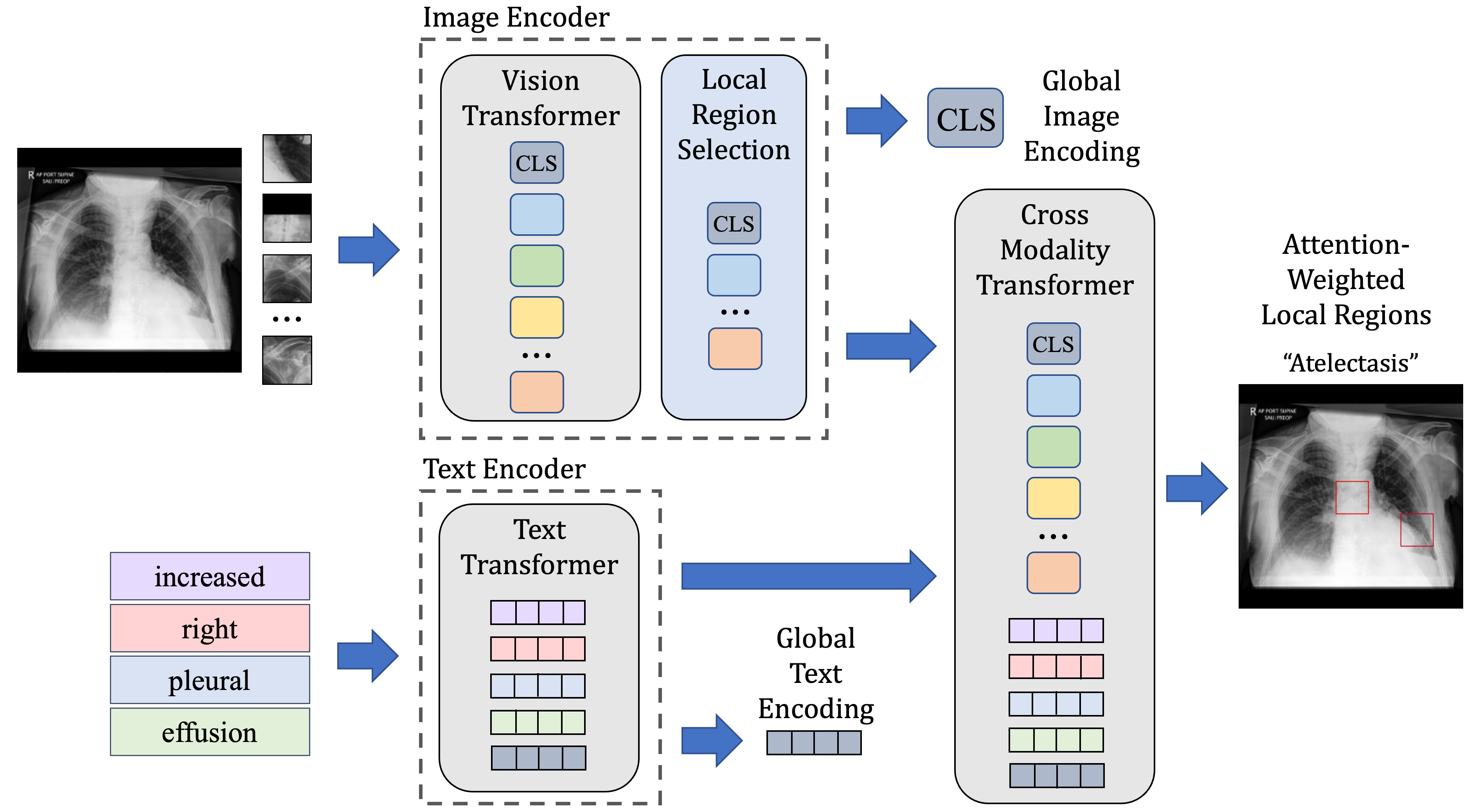}
 \caption{Overview of Local Contrastive Learning (LRCLR) framework.}
 \label{fig:pipeline_diagram_figure}
\end{figure}

\section*{3 \quad Method}

\textbf{Datasets}. Self-supervised fine-tuning was done on the MIMIC-CXR dataset, a public dataset of 377,100 chest radiographs drawn from 227,835 radiographic studies \cite{johnson2019mimic}. Each study comes with a corresponding radiology text report containing multiple sections written by radiologists: examination, indication, impression, findings, techniques, and comparison. We fine-tune the model for 20 epochs on the MIMIC-CXR dataset \cite{johnson2019mimic}. Image-text pairs are constructed by taking the chest radiograph and the impressions section of the corresponding radiology report. The zero-shot evaluation was done on the CheXpert dataset, a public chest X-ray dataset commonly used to train and evaluate the performance of radiology image classifiers \cite{irvin2019chexpert}. The CheXpert dataset consists of 224,316 chest radiographs taken from 65,240 patients, with a free-text radiology report associated with each radiograph. Each radiograph is also labeled for the presence of 14 different medical observations. Following \cite{tiu2022expert}, we evaluate the zero-shot performance of our models on the CheXpert test dataset, which consists of 500 chest radiographs.

\textbf{Image and Text Encoding}. An overview of the LRCLR framework is presented in Figure~\ref{fig:pipeline_diagram_figure}. The architecture consists of a Vision Transformer (ViT) \cite{dosovitskiy2020image} based image encoder and a transformer text encoder. We initialize our image encoder with the weights of a ViT pretrained on ImageNet \cite{russakovsky2015imagenet}, which takes as input a 224x224 resolution chest radiograph. The image encoder maps the input image into a global embedding which summarizes the information in the image. Local features are encoded in the patch embeddings, which capture information about subregions within the image. For the text encoder, we use a 12-layer transformer with a hidden dimension of 512 and 8 attention heads, following the parameters set in \cite{tiu2022expert}. The initialization scheme for the text encoder follows the original CLIP architecture, and a maximum context length of 77 tokens for the text extracted from the findings section of the radiology report. \cite{radford2021learning}.

\begin{figure}[t]
  \centering
\includegraphics[width=0.9\textwidth]{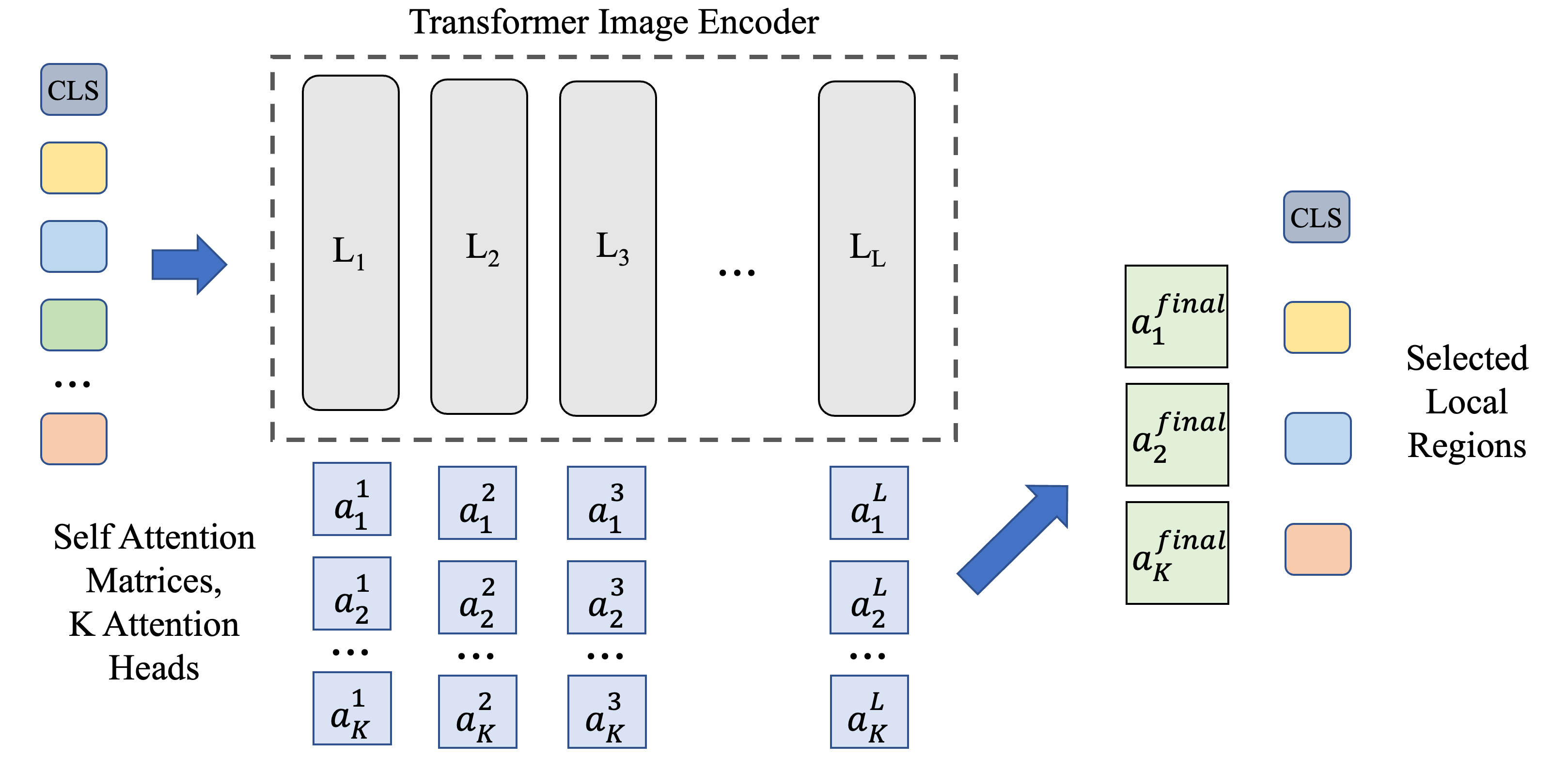}
 \caption{Overview of local region selection module.}
 \label{fig:local_region_selection_pipeline}
\end{figure}

\textbf{Local Region Selection}. Accurately selecting local image regions is critical for differentiating chest radiographs. Medical observations manifest themselves in small regions of chest x-rays, resulting in subtle variations between different chest radiographs. We draw inspiration from fine-grained visual recognition tasks in Computer Vision, which aim to classify subclasses with small inter-class variation within an object category, such as birds \cite{van2018inaturalist} or cars \cite{krause20133d}. Methods based on region-wise annotations are unfeasible given the scarcity and cost of annotating chest radiographs. Therefore, we turn to unsupervised methods for local region detection. Following \cite{he2022transfg}, we utilize the self-attention matrices learned as a part of the ViT image encoder to select important local regions. Attention matrices are calculated within each layer of the image encoder as a product of query and key vectors, which are obtained from the input sequence of vectors. Given $N$ input vectors $h_1^l, ..., h_N^l$ as a matrix $H^l$ for a given layer:

\begin{equation}
Q = H^l \cdot W_{query}^l
\end{equation}
\begin{equation}
K = H^l \cdot W_{key}^l
\end{equation}
\begin{equation}
a_l = softmax(\frac{QK^T}{\sqrt{d_k}})V
\end{equation}

Where $W_{query}^l$ and $W_{key}^l$ are learned projection matrices for query and key vectors, and $\sqrt{d_k}$ is the root of the hidden dimension of the query and key vectors. We extract self-attention matrices from all L layers and K heads of the image encoder, and perform matrix multiplication to obtain a representative matrix for each attention head:

\begin{equation}
a_k^{final} = {\displaystyle \prod_{l=0}^{L} a_k^{l}}
\end{equation}

where k is one of the K attention heads. $a_{final}$ captures how information about local regions propagates throughout the layers of the ViT, giving a better attention-based selection compared to a single layer of attention \cite{he2022transfg}. We take the maximum attended-to index for each attention head and select the corresponding K image regions. By keeping only selected image regions as well as the class token, we allow the model to focus on discriminative regions of images while also keeping the global image context.

\textbf{Cross-Modality Transformer}. Selecting local regions allows the model to focus on important local regions within an image. However, it cannot by itself provide any interpretation of the relationship between significant image regions and the accompanying radiology text. Additionally, the local regions may be underrepresented in the global image encoding, which contributes to the alignment between chest radiographs and radiology text. We therefore propose a cross-modality transformer module as a second add-on for fine-tuning, accepting as input a sequence of selected image region tokens and text tokens. A local contrastive loss \cite{radford2021learning, zhang2022contrastive} is calculated between the output image class token $v_l^i$ and text class token $t_l^i$ from the cross-modality layer:

\begin{equation}
L_{local} = \sum_{i=1}^{N} -log\left(\frac{e^{v_l^i \cdot t_l^i}}{\sum_{k=1}^N e^{v_l^i \cdot t_l^k}}\right)
\end{equation}

By jointly encoding image regions and text and aligning their representations through local contrastive learning, we emphasize the interaction between selected image regions and text. Our overall loss function can then be calculated as:

\begin{equation}
L = L_{global} + \lambda L_{local}
\end{equation}

Where $L_{local}$ is the global contrastive loss between the global image class token $v_g^i$ and global text class token $t_g^i$ output from the region selection module. Parameter $\lambda$ is left as a hyperparameter for weighing local loss relative to global loss. We use $\lambda = 0.5$ for our experiments, chosen through experimentation on our zero-shot task.

\newcolumntype{A}{>{\centering}p{0.1\textwidth}}

\begin{table}[H]
\begin{center}
\begin{tabular}{|l|A|A|A|A|c|}
\hline
Model     & Specificity  & Precision  & Recall  & F1-Score  & $\quad$ AUC $\quad$   \\
\hline
GLoRIA  & 61.81  & 28.13  & 65.48  & 34.18  & 55.39 \\
\hline
ConVIRT                 & 59.84  & 25.32  & \textbf{67.27}  & 32.56  & 58.64 \\
\hline
CheXzero                & \textbf{69.46}  & 40.79  & 65.2  & 45.72  & 76.24 \\
\hline
CheXzero + LRCLR         & 66.86  & \textbf{41.04}  & 64.25  & \textbf{46.79}  & \textbf{78.66} \\
\hline
\end{tabular}
\end{center}
\caption{Summary of zero-shot classification performance on the CheXpert test set. Metrics are calculated over all 14 CheXpert findings.}
\label{tab:zero_shot_results_table}
\end{table}


\begin{figure}[H]
  \centering
    \includegraphics[width=0.9\textwidth]{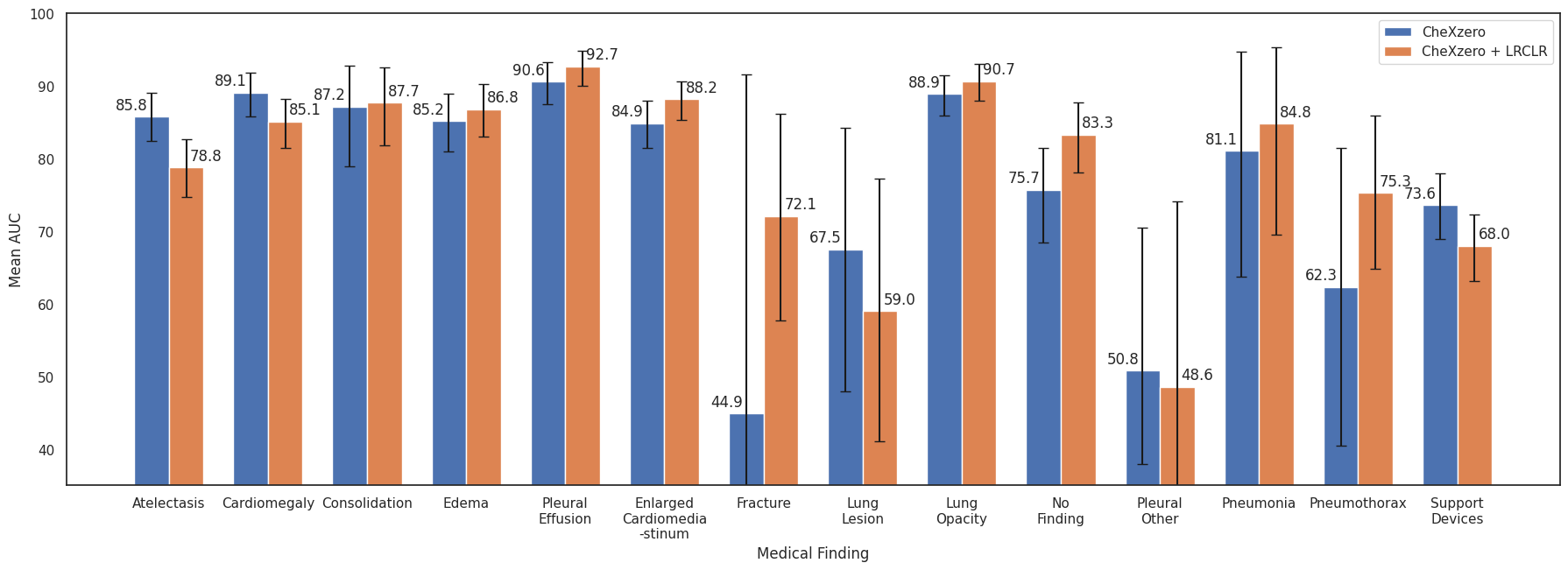}
 \caption{AUC performance across all 14 CheXpert medical observations.}
 \label{fig:auc_barplot}
\end{figure}

\section*{4 \quad Experiments \& Results}

\textbf{Zero-Shot Classification}. We evaluate the effectiveness of our proposed fine-tuning module in a zero-shot classification setting on the CheXpert test dataset following the methodology outlined in Section 3. We first take image-text pairs of the chest x-rays and zero-shot classification labels associated with the CheXzero dataset, and encode them using the fine-tuned model. The encoded global image and text representations are then contrasted against each other to determine which text labels the image most corresponds with. Additionally, we extract attention weights from the cross-modality transformer module to inspect the relationship between selected image regions and zero-shot labels. To evaluate the performance of baseline models and our proposed method, predictions are obtained from each model on the CheXpert test dataset, and metrics are calculated for each of the 14 findings. We report the average specificity, precision, recall, F1-score, and AUROC for each of the models across all 14 medical observations. For CheXzero \cite{tiu2022expert}, we use the pretrained CLIP checkpoint from \cite{radford2021learning}, and fine-tune on the MIMIC-CXR dataset \cite{johnson2019mimic}. For the GLoRIA \cite{huang2021gloria} baselines, we load the publicly available model and perform zero-shot evaluation on the CheXpert test dataset. For ConVIRT, since there is no publicly available code, we implement the method according to \cite{zhang2022contrastive} and evaluate it on the CheXpert dataset.

\begin{figure}[t]
  \centering
    \includegraphics[width=0.8\textwidth]{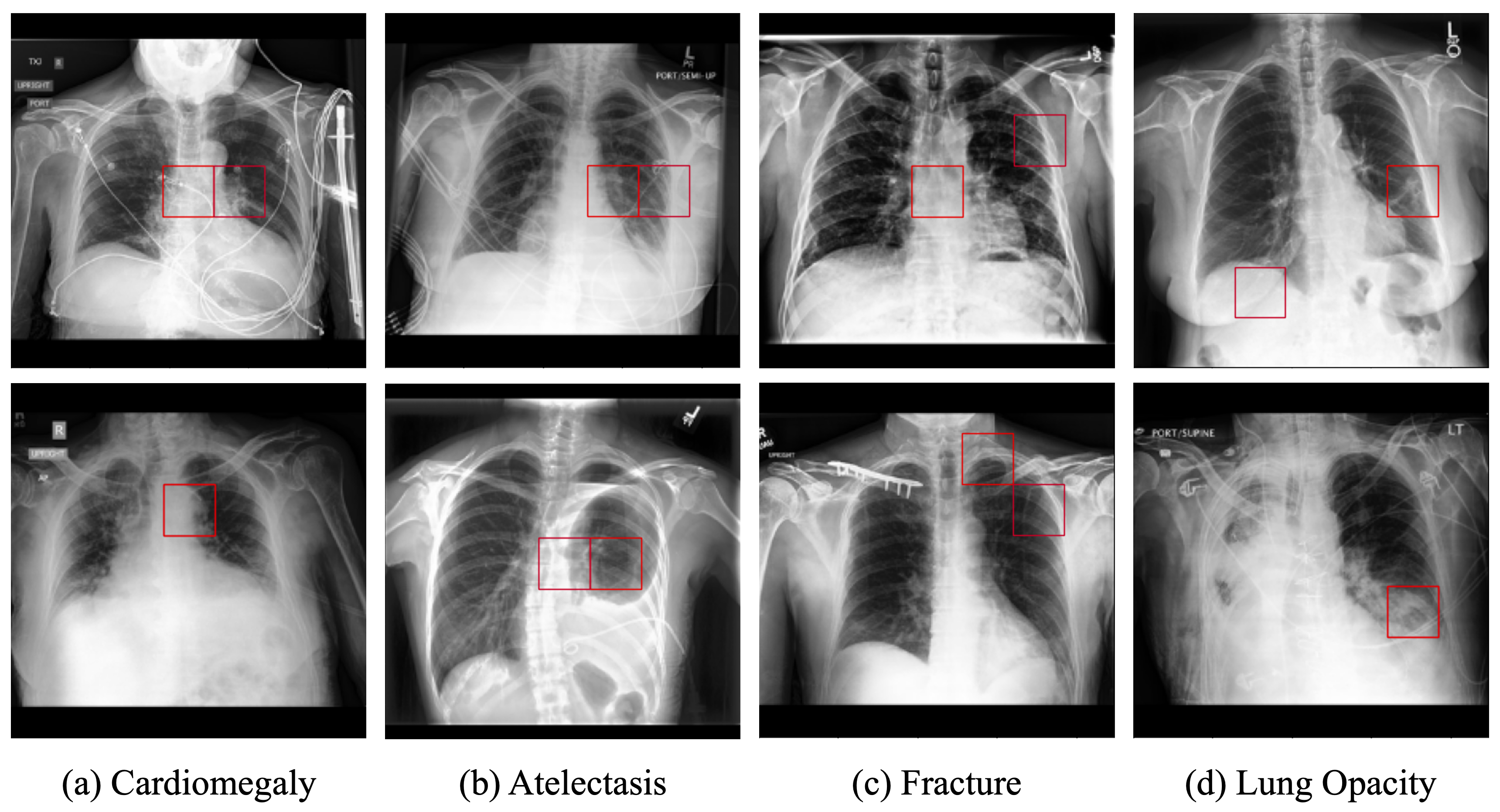}
 \caption{Examples of chest radiographs with top 2 selected regions highlighted. A single highlighted region indicates that two attention heads highlighted the same image region with a high score.}
 \label{fig:regionwise_interpretability}
\end{figure}

We present the comparison of our proposed fine-tuning module against baseline methods in Table~\ref{tab:zero_shot_results_table}. Based on our experimental results, we observe that the addition of our fine-tuning module led to equal or improved performance in three of the five evaluated metrics, including a 2.42 improvement in AUROC across all 14 medical observations in the CheXpert test dataset. A comparison of AUROC across each of the 14 findings (Figure \ref{fig:auc_barplot}) shows that the region selection and local contrastive learning resulted in improved performance on seven of the medical observations, most notably a 27.2 increase in AUC on Fracture observations, a 7.6 increase on No Finding observations, and a 13.0 increase on Pneumothorax observations.

From these results, we infer that our local region selection module is effectively selecting significant image regions in the chest radiographs, allowing the model to focus on more discriminative regions between different images. We note that the number of selected local regions corresponds to the number of attention heads in the image encoder, and that two attention heads may attend highly to the same image region.

In Figure~\ref{fig:regionwise_interpretability}, we visualize the attention weights between selected local regions and zero-shot labels extracted from the cross-modality transformer module. The highlighted regions indicate significant image regions selected by the region selection module, colored by attention score from lowest (dark blue) to highest (dark red). We observe that the attention scores and selected regions correctly identify significant image regions corresponding to the text prompt, and provide an ordering when comparing image regions to a given text prompt. We emphasize that, unlike previous works, our fine-tuning modules are compatible with many contrastive attention-based architectures, and can be added in for additional interpretability across image regions and text. Previous works \cite{jain2019attention} have studied the interpretability of attention coefficients in deep networks, and while the scores may not provide complete interpretation, a well-trained contrastive model can still highlight significant image regions corresponding to a text prompt.

\begin{figure}[t]
  \centering
    \includegraphics[width=0.8\textwidth]{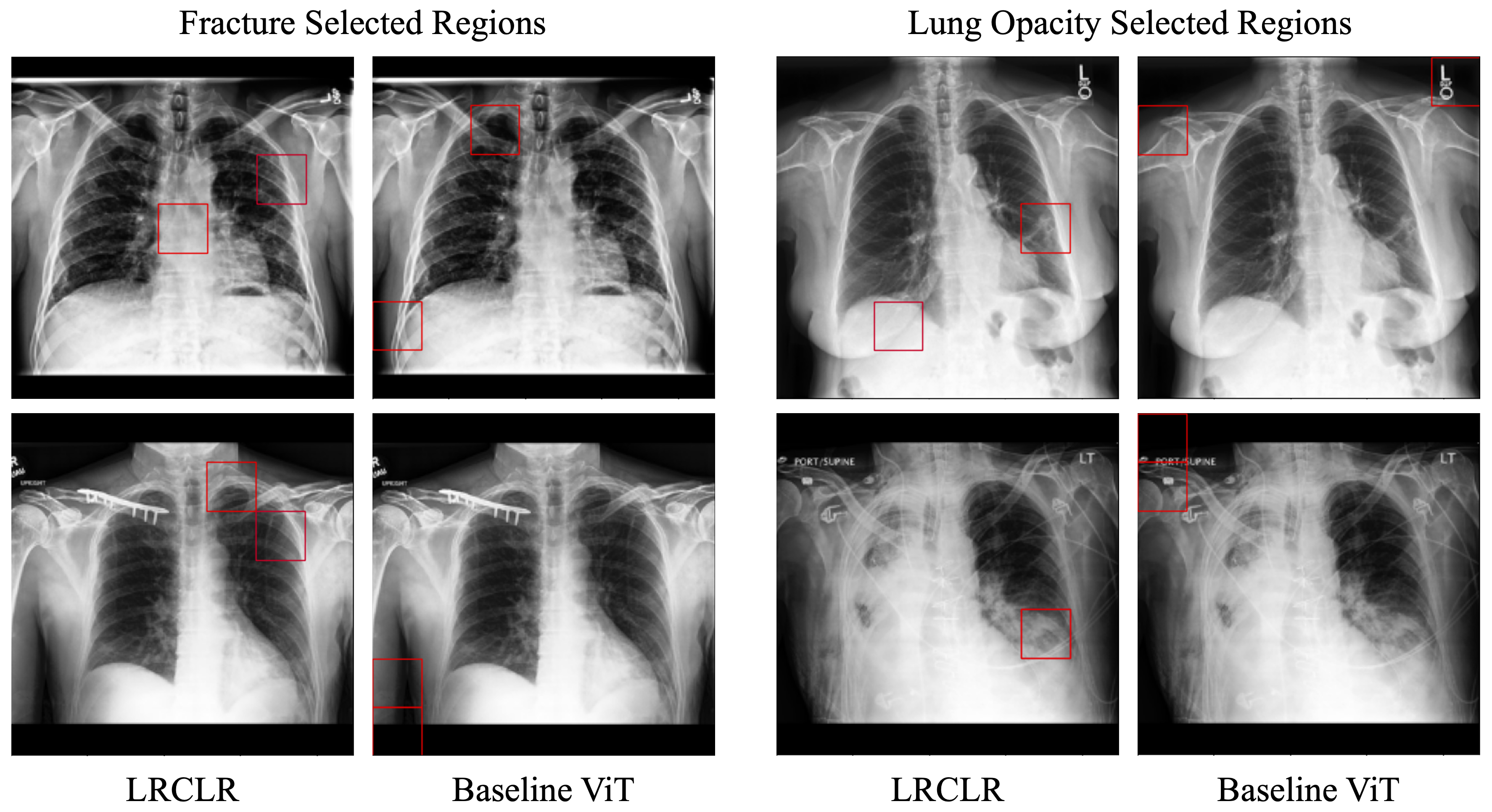}
 \caption{Comparison of selected regions from the LRCLR framework and baseline CheXzero model. Baseline selected regions were obtained by extracting attention weights from the ViT image encoder and following the same procedure as outlined in Section 3.}
 \label{fig:case_study_selected_regions}
\end{figure}

\textbf{Case Study: Fracture}. We conducted an analysis of our fine-tuning module on the fracture medical observation in the CheXpert test set. We identified the largest improvement in AUROC over baseline methods on this finding. Attention visualizations, presented in Figure \ref{fig:regionwise_interpretability}, revealed that the trained model often focused on the chest rib regions in chest radiographs - a common site for fractures. Such rib fractures can result in additional complications for patients \cite{nazal2012rib}, and chest radiography is a reliable method for their diagnosis. Our computational aid, the cross-modality module, can provide valuable interpretability between significant regions and text, and aid radiologists in identifying critical regions that require further analysis. 

We compare our selected image regions to baseline selected regions in Figure \ref{fig:case_study_selected_regions}. Baseline selected regions were obtained by extracting attention weights from the Vision Transformer image encoder in the baseline CheXzero model, and following the procedure outlined in Section 3 to obtain highlighted regions per attention head. We rank the baseline selected regions by the associated attention score from the corresponding attention head, and choose the top two scored image regions for comparison to our method. We note that baseline selected regions do not effectively highlight the rib region in the chest radiographs, and instead focus on other outer portions of the image.

\textbf{Case Study: Lung Opacity}. We additionally evaluate our fine-tuning module on the lung opacity medical observation in the CheXpert test set, which also saw improved performance compared to baseline zero-shot performance. Attention visualizations from our method \ref{fig:regionwise_interpretability} highlight regions in the lower lung, where uncertain areas of white material are present in the radiograph. These regions may be indications of lung complications. In comparison to baseline selected regions shown in Figure \ref{fig:case_study_selected_regions}, we observe that our method focuses more on unclear lung regions, whereas the baseline highlights regions on the edges of the radiograph.

While insightful, we acknowledge that the selected regions highlighted by our method do not always correspond well to the intended text prompt. Highlighted regions may focus on different regions within the radiograph, particularly in radiographs with many conditions present or in noisy images. We do, however, note that by ranking image regions with attention corresponding to the given text prompt, we can filter out some of these uninsightful selected regions.

\subsection*{6 \quad Related Works}

In this section, we review related literature on contrastive learning methods and local representational learning.

\textbf{Contrastive vision-language frameworks}. CLIP \cite{radford2021learning} extracts vision and language embeddings using image and text encoders which are trained by contrastive learning by a symmetric cross-entropy loss. Similarity scores are calculated between image and text embeddings, which are used after pretraining for downstream tasks such as zero-shot classification and image-to-text retrieval. ConVIRT \cite{zhang2022contrastive} utilized the idea of contrastive learning to pretrain on medical image-text pairs. MedCLIP \cite{wang-etal-2022-medclip} decouples paired images and texts and uses soft targets of semantic similarities to learn from unpaired medical images and text. CheXzero \cite{tiu2022expert} is a direct application of the CLIP model on large-scale chest X-ray datasets to enable zero-shot classification of unseen findings in images. Seibold et al. \cite{seibold2022breaking} improved the CLIP model by adding a local contrastive loss function that contrasts local image features to sentence features in the radiology reports. LoVT \cite{muller2022joint} extracts localized representations by the local alignment between output feature maps (image sub-region features) and max pooling of token embeddings (sentence embeddings). BioViL \cite{boecking2022making} develops a new vision-and-language pre-training model and proposes a phrase grounding task to evaluate the local alignments between images and texts. MGCA \cite{wang2022multigranularity} further uses three levels of embedding alignments, i.e., pathological region, instance, and disease, to combine local, global, and cluster representations for downstream tasks.

\textbf{Global-Local Representational Learning}. Chaitanya et al. proposed a framework \cite{chaitanya2020contrastive} that uses self-supervised learning (SSL) and contrastive learning for semi-supervised segmentation of volumetric medical images with limited annotations. They leverage structural similarity across images and a local contrastive loss to learn distinctive local representations for per-pixel segmentation. GLoRIA \cite{huang2021gloria} learns a global image and text embedding, and additionally learns attention weights between individual region and word embeddings through a local contrastive objective. Although similar to our work, our proposed method differs from GLoRIA in that we perform self attention-based local region selection to discard unimportant local regions before the cross-modal transformer module and local contrastive loss. Additionally, we formulate our framework as a flexible fine-tuning module, which can be added to any contrastive-based architecture for added interpretability.

\subsection*{7 \quad Conclusion}

We introduce a flexible fine-tuning module for contrastive transformer-based frameworks, designed to enhance cross-modal interpretability in chest radiograph analysis tasks. Our two-part module, named LRCLR, comprises a region selection layer and a cross-modal interaction layer, offering interpretable scores between image regions and associated text. Visualizations of attention scores and selected regions exhibit the model's ability to identify and emphasize crucial image regions, thereby enabling cross-modal interpretability for radiologists analyzing chest radiographs. Our work highlights the versatility of fine-tuning to incorporate interpretability in contrastive architectures. We suggest extending local region selection to non-attention-based image encoders and exploring various unsupervised region selection procedures for attention-based methods to improve this approach.

\section*{Acknowledgement}
XJ is CPRIT Scholar in Cancer Research (RR180012), and he was supported in part by Christopher Sarofim Family Professorship, UT Stars award, UTHealth startup, the National Institute of Health (NIH) under award number R01AG066749, R01LM013712, and U01TR002062, and the National Science Foundation (NSF) \#2124789

\pagebreak

\makeatletter
\renewcommand{\@biblabel}[1]{\hfill #1.}
\makeatother

\bibliographystyle{vancouver}
\bibliography{amia}  

\end{document}